\documentclass[10pt,twocolumn,letterpaper]{article}

\usepackage{cvpr}              

\usepackage[dvipsnames]{xcolor}

\definecolor{cvprblue}{rgb}{0.21,0.49,0.74}
\usepackage[pagebackref,breaklinks,colorlinks,citecolor=cvprblue]{hyperref}
\usepackage{wrapfig}

\usepackage{multirow}

\def\paperID{73} 
\def\confName{3DV\xspace}
\def\confYear{2025\xspace}

\title{Direct and Explicit 3D Generation from a Single Image}

\author{
Haoyu Wu\textsuperscript{1*} \qquad
Meher Gitika Karumuri\textsuperscript{2} \qquad
Chuhang Zou\textsuperscript{2} \qquad 
Seungbae Bang\textsuperscript{2}\\
Yuelong Li \textsuperscript{2} \qquad
Dimitris Samaras \textsuperscript{1} \qquad
Sunil Hadap \textsuperscript{2}\\
\textsuperscript{1}Stony Brook University \qquad
\textsuperscript{2}Amazon Inc.
}

\begin{document}
\maketitle
\let\thefootnote\relax\footnotetext{$^\ast$ This work was done as part of Haoyu's internship at Amazon.}

 \begin{abstract}

Current image-to-3D approaches suffer from high computational costs and lack scalability for high-resolution outputs. 
In contrast, we introduce a novel framework to directly generate explicit surface geometry and texture using multi-view 2D depth and RGB images along with 3D Gaussian features using a repurposed Stable Diffusion model. 
We introduce a depth branch into U-Net for efficient and high quality multi-view, cross-domain generation and incorporate epipolar attention into the latent-to-pixel decoder for pixel-level multi-view consistency. 
By back-projecting the generated depth pixels into 3D space, we create a structured 3D representation that can be either rendered via Gaussian splatting or extracted to high-quality meshes, thereby leveraging additional novel view synthesis loss to further improve our performance. 
Extensive experiments demonstrate that our method surpasses existing baselines in geometry and texture quality while achieving significantly faster generation time.

\end{abstract}    
 \section{Introduction}
\label{sec:intro}

\begin{figure}[!htb]
\begin{center}
\centerline{\includegraphics[width=0.5\textwidth]{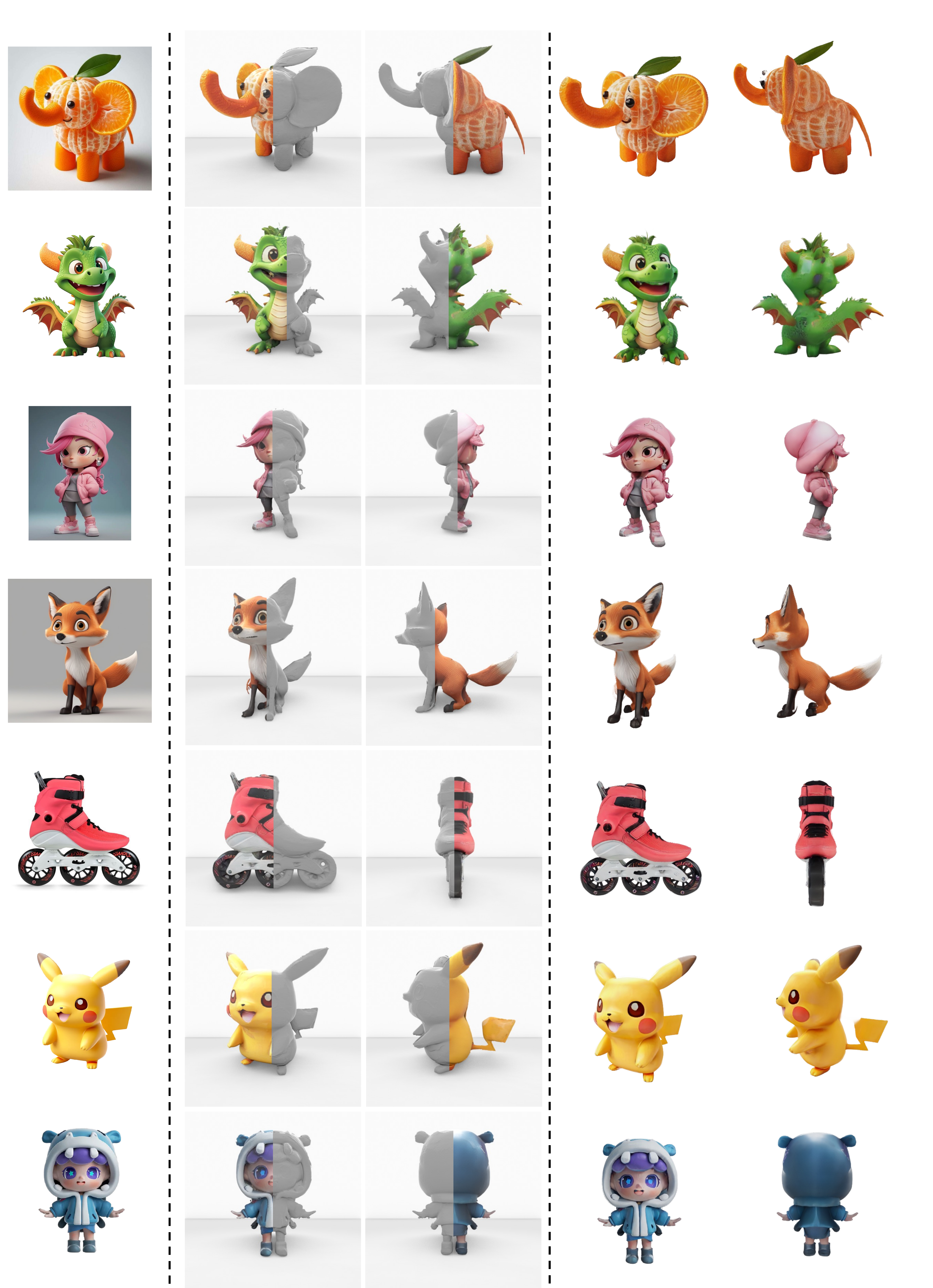}}
\end{center}
\vspace{-25pt}
\caption{We present our approach that generates high-resolution (x512), textured 3D asset from a single image. From left to right: input in-the-wild images downloaded from the internet, generated 3D textured meshes, novel view synthesis via Gaussian splatting.}
\label{fig: fig1}
\vspace{-10pt}
\end{figure}

The task of generating 3D assets from single images \cite{jun2023shap, liu2023zero, liu2023syncdreamer, melas2023realfusion, nichol2022point, qian2023magic123} is pivotal in several application domains, such as 3D content creation, virtual reality, augmented reality, as well as 3D aware image generation and editing. However, 
building a 3D model from a sparse set of images, let alone just one, is a highly ill-posed problem. There are inherent ambiguities in this inverse rendering problem, 
and the greatest challenge is effectively ``hallucinating'' unseen portions of the object in terms of geometry and texture. Recent cutting-edge generative AI approaches (e.g. diffusion models, transformers) attempt to overcome these obstacles by learning powerful 3D priors and show promising results. 

\begin{figure*}[tb]
\begin{center}
\centerline{\includegraphics[width=0.99\textwidth]{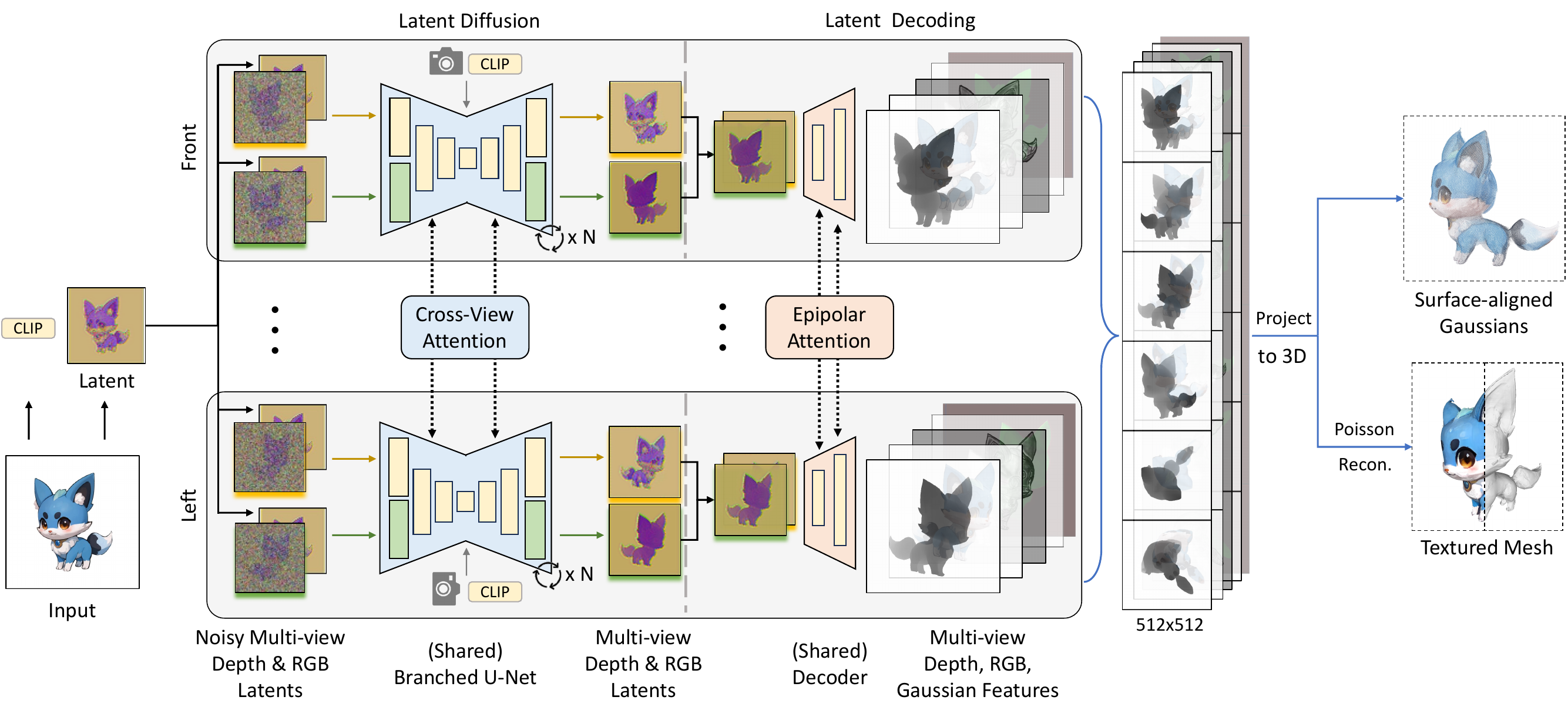}}
\end{center}
\vspace{-30pt}
\caption{\textbf{Overview}. 
Our method is a feed-forward image-to-3D model. Given an input image, we generate depth and RGB latent images from six orthographic views via simultaneous multi-view diffusion. The process is conditioned on input latent, input CLIP embedding, and cameras. We incorporate a branched U-Net for efficient and high-quality cross-domain diffusion. For each view, we channel-concatenate depth and RGB latents and decode it to depth, RGB, and Gaussian features in pixel space (512x512 resolution). We add epipolar attention in the decoder, which is crucial for generating pixel-level multi-view consistent depths. We lift our output (RGB and opacity from Gaussians) into 3D space via depth unprojection, creating high-quality textured mesh via Poisson surface reconstruction. Additionally, our lifted surface-aligned 3D Gaussians enable novel view synthesis via Gaussian splatting, allowing additional gradient decent loss from NVS.}
\label{fig: main}
\vspace{-10pt}
\end{figure*}

A major technical challenge for 3D generation is how to represent 3D objects / scenes that can be easily modeled. Recent techniques seek for volumetric representation, such as Neural Radiance Fields (NeRF) \cite{mildenhall2021nerf, jun2023shap} or triplane \cite{chan2022efficient, gupta20233dgen}. 
Such volumetric representation typically has high computational and memory complexity which inhibits its scalability towards high-quality and high-resolution generation.
In pursuit of efficient explicit representation, recent works directly generate point clouds~\cite{nichol2022point, vahdat2022lion, luo2021diffusion, zou2024triplane} or meshes~\cite{liu2023meshdiffusion}. Although explicit generations significantly reduce computational complexity compared to volumetric representations, they represent 3D quantities in a domain which significantly deviates from that of natural images, posing great challenges in borrowing strong 2D priors from contemporary foundation models such as Stable Diffusion (SD)~\cite{rombach2022high}.
Recently, an alternative two-stage strategy \cite{liu2023syncdreamer, long2024wonder3d, gao2024cat3d} was proposed that first generates multi-view images and subsequently fits a 3D representation out of them. However, effectively enforcing consistency across multiple views remains a challenge. Methods that only generate RGB images \cite{gao2024cat3d, liu2023syncdreamer} typically require dense views to form extensive coverage of the scene/object, which can be computationally expensive.

To address the aforementioned challenges, we propose to directly generate explicit surface geometry and texture using multi-view 2D RGB, depth and Gaussian feature images. We believe that this representation offers a more scalable approach towards high-resolution and detail-preserving generation, and we hypothesize that depth maps capture geometry information more effectively than other alternatives like normals. 
We repurpose the Stable Diffusion \cite{lambda_stable_diffusion, rombach2022high} model and additionally introduce a branched U-Net with expert blocks for efficient and high quality multi-view image generation.
To ensure pixel-level multi-view consistent depth maps, we incorporate epipolar attention \cite{yang2022mvs2d, he2020epipolar, wang2022mvster} in the latent decoding process.
Our representation is versatile and can be easily converted to other formats.
In particular, we immediately obtain a structured 3D representation when we back-project multi-view depth, RGB images and additional Gaussian features to 3D, \ie dense surface-aligned point cloud or Gaussians. This 3D representation can be either extracted to high-quality textured mesh via Poisson surface reconstruction \cite{kazhdan2013screened} or rendered efficiently using Gaussian Splatting \cite{kerbl20233d} for novel view synthesis (NVS), thereby leveraging additional NVS loss to further improve our performance

In summary, our main \textbf{contributions} are:
\begin{itemize}
\item  We propose a novel streamlined framework to predict multi-view depth maps along with RGB and Gaussian features by fine-tuning a pre-trained 2D diffusion model. Our representation offers a compact encoding of the 3D models and scales well towards high-resolution generation, enabling high-fidelity content creation.
\item We develop a branched U-Net that learns to efficiently generate multi-view RGB and depth images while leveraging cross-domain similarities.
\item To enforce pixel-level multi-view consistent depths, we incorporate epipolar attention in latent-to-pixel decoder following geometric insights.
\item We demonstrate significant improvements in quality and speed over existing benchmarks, marking a major advancement in single-view 3D generation.
\end{itemize}
 
\section{Related Work}
\label{sec:related_work}



\noindent \textbf{Representation of 3D models.}
A fundamental issue to address when formulating 3D generation is how to represent the 3D model, which can be either explicit surface-like representation or implicit volumetric representation. Typical examples of explicit representation include Point-E \cite{nichol2022point} and LION \cite{vahdat2022lion} that train diffusion models to generate point clouds, or MeshDiffusion \cite{liu2023meshdiffusion} that generates parametrized 3D meshes. One drawback of directly generating explicit representation is that it becomes difficult to reuse 2D image priors due to the domain gap between images and explicit representation formats such as point clouds or meshes. 
In contrast, our method predicts depth images which enable a shared 2D latent representation and facilitate incorporation of 2D priors.

Since the popularity of NeRF \cite{mildenhall2021nerf}, many recent techniques adopt volumetric representation such as neural fields \cite{jun2023shap, anciukevivcius2023renderdiffusion, chen2023single, cheng2023sdfusion, erkocc2023hyperdiffusion, gu2024control3diff, gupta20233dgen, karnewar2023holofusion, kim2023neuralfield, muller2023diffrf, ntavelis2024autodecoding, wang2023rodin, zhang20233dshape2vecset}, triplane \cite{chan2022efficient,hong2023lrm,he2023openlrm,wang2024crm,xu2024instantmesh,zou2024triplane}, etc. However, these approaches typically require expensive volumetric rendering, posing great challenges on scaling up to high-resolution predictions.
For example, LRM \cite{hong2023lrm} has triplane resolution of just 64x64, relying on low resolution features to render high resolution images. 
In contrast, our method predicts multiple sparse views which naturally scale towards high-resolution.


Some recent feed-forward methods \cite{zhang2024gs, xu2024agg, tang2024lgm, zou2024triplane, szymanowicz2023splatter, xu2024grm} also propose to generate Gaussians as 3D representation. However, our approach primarily focuses on generating pixel-perfect multi-view depth maps, leading to significantly improved generation quality.

\noindent \textbf{Generation with Score Distillation Sampling.}
Generation of 3D objects / scenes from single or sparse input views is an inherently ambiguous problem. Recent advancements in 3D generation endeavored to leverage large 2D generative models \cite{ramesh2021zero, saharia2022photorealistic, rombach2022high} as strong image priors by reformulating the problem into 2D domains. These 2D generative models are trained on extensive internet-scale image datasets, which generalizes over diverse scenarios. A notable innovation in this area has been introduced by DreamFusion \cite{poole2022dreamfusion} which introduced Score Distillation Sampling (SDS).  The core methodology involves optimizing a parameterized 3D representation, \eg  NeRF \cite{mildenhall2021nerf}, SDF, or mesh using a pre-trained 2D prior model to supervise rendered views. This technique has been later successfully applied to both text-to-3D and image-to-3D synthesis tasks \cite{armandpour2023re, chen2023fantasia3d, chen2024it3d, huang2023dreamtime, lin2023magic3d, seo2023ditto, seo2023let, tsalicoglou2024textmesh, wang2024prolificdreamer, wu2024hd, yu2023points, zhu2023hifa, qiu2024richdreamer, wang2023imagedream}, demonstrating its versatility and effectiveness. Despite the early promise of this approach, the generated 3D objects often lack 3D consistency. 
In addition, it requires a time-consuming per-scene optimization process which limits its application in real-time scenarios.

\noindent \textbf{Generation and Fusion of Multiple Views.}
Another research direction focuses on generating multi-view images from a single image \cite{liu2023zero, chan2023generative, deng2023nerdi, gu2023nerfdiff, lei2022generative, liu2023deceptive, szymanowicz2023viewset, tewari2024diffusion, tseng2023consistent, watson2022novel, xiang20233d, yoo2024dreamsparse, yu2023long, zhou2023sparsefusion, gao2024cat3d, voleti2024sv3d}. Zero123 \cite{liu2023zero} pioneers in adapting the pre-trained 2D diffusion model for multi-view image synthesis by incorporating camera conditions into the model. While this approach delivers promising results, it lacks constraints across different views and hence struggles with consistency across the generated multi-view images. To mitigate this, SyncDreamer \cite{liu2023syncdreamer} introduces a volume-based multi-view information aggregation module using 3D CNN and spatial attention. One-2-3-45 \cite{liu2024one} aims to combine 2D generative models and multi-view 3D reconstruction. Some other works \cite{shi2023mvdream, long2024wonder3d} leverage 2D dense cross-view attention to enhance 3D consistency. With the generated multi-view images available, these methods then fit a 3D representation via the reconstruction loss. While these methods offer qualitative improvement and an alternative path to SDS, they still struggle with multi-view consistency and may fail to produce high-quality meshes. 

In addition, many such techniques only predict RGB images and thus require abundant views to form a dense coverage of the 3D models. Therefore, the subsequent optimization process can be time-consuming, typically lasting at least several minutes per scene. Among recent attempts \cite{stan2023ldm3d, ke2024repurposing} to embrace depth prediction, MVD-Fusion \cite{hu2024mvd} was proposed that leverages intermediate low-resolution depth maps as fusion guidance. However, it does not produce high-resolution depth maps as end predictions, which we believe is important for high-quality 3D reconstruction.

Other methods explore alternative 3D related images \cite{li2023sweetdreamer, liu2024pi3d} including normals \cite{long2024wonder3d}. For normal maps, we find it is difficult to derive depth discontinuities out of normal values, and 3D reconstruction from multiple normal maps typically require a dedicated optimization process. In contrast, depth maps faithfully capture geometric details, enabling high-fidelity reproduction of object shapes.



 \section{Preliminaries}

\noindent \textbf{3D Gaussian splatting} performs novel view synthesis using Gaussians as a 3D representation. Each Gaussian in 3D is defined by a center $\mathbf{x} \in \mathbb{R}^3$, a color feature $\mathbf{c} \in \mathbb{R}^C$, an opacity value $\alpha \in \mathbb{R}$, a scaling factor $\mathbf{s} \in \mathbb{R}^3$, and a rotation quaternion $\mathbf{q} \in \mathbb{R}^4$. Renderings are performed via alpha composition using differentiable rasterization in real time. 

\noindent \textbf{Pixel-aligned 3D Gaussians.} The Gaussian center $\mathbf{x}$ can be replaced by depth pixels $\mathbf{t} \in \mathbb{R}$. Suppose that $\text{ray}_o$ and $\text{ray}_d$ represent the ray origin and ray direction, respectively; the Gaussian center is then inferred as $\mathbf{x} = \text{ray}_o + \mathbf{t} \cdot \text{ray}_d$. 
The final output can be obtained by merging the 3D Gaussians from $N$ views, resulting in $N \cdot HW$ Gaussians.

\section{Method}
\label{sec:method}

Given an input image, our goal is to generate the 3D shape with high-resolution textures and high-fidelity geometry. We propose to directly generate explicit surface geometry by decomposing the 3D representation into multi-view consistent outputs, each of which contains a depth map, an RGB image, and a Gaussian feature map. 
Predictions from different views are used to construct a 3D point cloud, which can be extracted into a textured mesh or rendered using 3D Gaussian splatting~\cite{kerbl20233d}, enabling natural support for novel view synthesis (NVS) and taking advantage of additional NVS loss to further improve our performance.

An overview of our method is illustrated in~\cref{fig: main}. 

\subsection{Multi-view Generation in Latent Space}
Reconstructing 3D objects from one single input view is an ill-posed problem in general. Injecting prior knowledge is critical to resolving the inherent ambiguity. To this end, we re-purpose the Stable Diffusion model~\cite{rombach2022high}, which possesses rich domain knowledge through internet-scale pre-training, to predict multi-view latents. Since Stable Diffusion generates each view independently, it could potentially generate inconsistent views, causing blurriness or distortions in the generated 3D objects. We follow prior works \cite{liu2023syncdreamer, shi2023mvdream, long2024wonder3d} to extend the self-attention layer in the Stable Diffusion U-Net into dense cross-attention among different views, so that features are not only attended spatially but also across views. Thus, we obtain multi-view consistent generation in latent space.

\subsection{Enforcing Cross-view Depth Consistency with Epipolar Attention}\label{ssec:method_epipolar}
Using the pre-trained VAE decoder in the Stable Diffusion framework, we decode multi-view latents back to multi-view RGB images. Although the dense cross-view attention is done in the latent space, we and prior works \cite{long2024wonder3d, shi2023mvdream, liu2023syncdreamer} also observe good consistency for the decoded multi-view RGB images.

Thus, a natural solution to predict multi-view depth maps is to follow the same framework: one could simply finetune the decoder to decode the multi-view latents back to multi-view depth maps. However, as shown in \cref{fig: abl} and \cref{tab: ablation} ("w/o Epipolar Attention"), we find this naive approach fails to produce 3D consistent depth maps, resulting in bad mesh extractions. The reason is that we use decoded images for explicit geometry representation, which requires higher pixel-level value accuracy and 3D consistency. This suggests dense cross-view attention in the latent space achieves consistency in RGB images but still falls short to estimate pixel-level accurate and multi-view consistent depth maps.

%

\begin{figure}[t]
\begin{center}
\centerline{\includegraphics[width=0.25\textwidth]{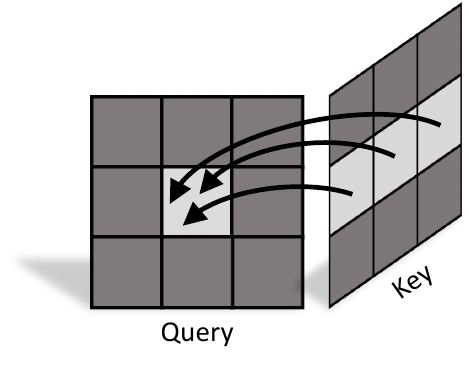}}
\end{center}
\vspace{-30pt}
\caption{An illustration of the epipolar attention in the decoder. We utilize epipolar geometry to facilitate efficient multi-view information exchange among views.}
\label{fig: method_epipolar}
\vspace{-10pt}
\end{figure}

To address this problem, we propose adding \emph{epipolar attention} \cite{yang2022mvs2d, he2020epipolar, wang2022mvster, kant2024spad, wang2023mvdd}, which is much more computationally efficient than the cross-view attention used in U-Net. It borrows insights from multi-view geometry, which dictates that corresponding pixels from different views are constrained to lie on certain lines called epipolar lines~\cite{hartley2003multiple}. In implementation, we modify the attention transformer component to reduce the number of computations, as depicted in~\cref{fig: method_epipolar}. Each token in the query image only attends to tokens from other views along the epipolar line. 
Since our approach predicts six othogonal views of front, back, left, right, top and bottom, the epipolar attention can be simplified as efficient row or column attention \cite{li2024era3d} because any pair of our output views is either parallel or perpendicular to each other.

\subsection{Efficient Cross-domain Denoising via Expert Branch}
\label{ssec:method_stable_diffusion}
Although epipolar attention greatly improves 3D consistency for the decoded depth maps, we find that the resulting depth maps lack details, as shown in \cref{fig: abl} and \cref{tab: ablation} ("w/o Depth Latent"). This suggests that relying solely on the RGB latent for decoding both RGB and depth, is insufficient for producing detailed depth maps. To mitigate this problem, 
we modify the latent U-Net to also output latent depth images along with latent RGB images. In practice, we find that channel-concatenating RGB and depth latents and feeding them to the decoder produces high-quality RGB and depth maps.

\begin{figure}[t]
\begin{center}
\centerline{\includegraphics[width=0.45\textwidth]{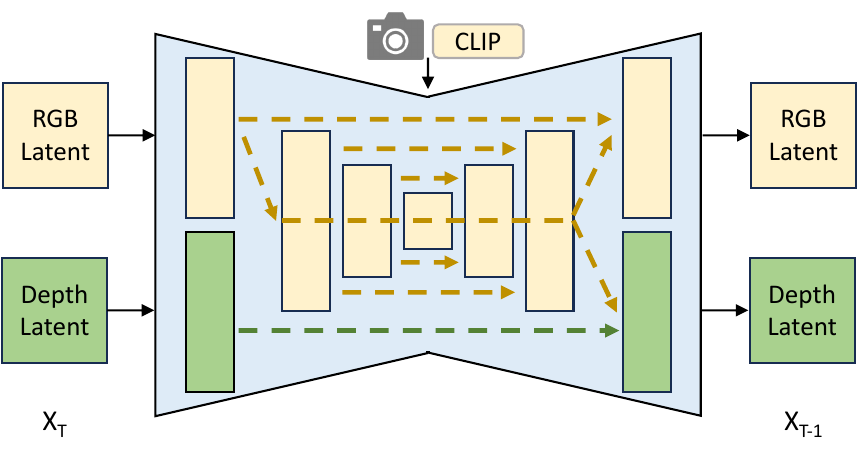}}
\end{center}
\vspace{-30pt}
\caption{An illustration of the branched U-Net for cross-domain latent diffusion. We add two blocks (green) as the depth branch for latent depth prediction. By sharing most weights with the original model that predicts RGB latents, we achieve efficient and high-quality depth prediction in a single inference.}
\label{fig: method_unet}
\vspace{-10pt}
\end{figure}

We observe that following Wonder3D's \cite{long2024wonder3d} architecture of using domain switch to infer RGB and depth images, doubles the required diffusion inference resources by employing separate labels for RGB and depth. As this process is time and memory intensive, instead, we implement a modified version of expert branch strategy proposed in HyperHuman \cite{liu2023hyperhuman} where we train for both RGB and depth latents simultaneously instead of successively. This improves the training and inference time significantly. As demonstrated in \cref{fig: method_unet}, we modify the U-Net so that each domain (RGB and depth) has its dedicated expert branch for the first DownBlock and the last UpBlock. We input noisy RGB and depth latents into their respective first DownBlocks. Then, the extracted features from RGB domain traverse the U-Net middle layers. These RGB features are then fed into the last UpBlock of different domain expert branches along with residual features from respective first DownBlocks. This produces individual branch outputs, \ie denoised RGB and depth latent. Additionally, we find this individual emphasis on RGB features effectively prevents catastrophic model forgetting in training. After obtaining RGB and depth latents, we channel-concatenate them together and forward it to the decoder.

\subsection{Efficient Rendering with 3D Gaussian Splatting}
We find that when we extend the decoder to predict additional Gaussian features and assign them to our colored point cloud, we essentially create pixel-aligned 3D Gaussians~\cite{zhang2024gs, chen2024mvsplat, szymanowicz2023splatter, charatan2024pixelsplat, wewer2024latentsplat, xu2024grm, tang2024lgm} that are also surface-aligned due to our multi-view consistent depth design. We find this simple decoder extension enables high-quality novel view synthesis through Gaussian splatting~\cite{kerbl20233d}. Thus, our decoder training can benefits from additional supervision via novel view synthesis loss, enhancing the performance.

\noindent \textbf{Predicting pixel-aligned Gaussians.} 
Using the generated RGB and depth images, we can create a dense colored point cloud by back-projecting all pixels into 3D space. We choose to further expand the output of the decoder to include additional 8 channels (1-channel opacity, 3-channel scale, and 4-channel rotation quaternion). Because we produce surface-aligned Gaussians \cite{guedon2024sugar}, we restrict their scales to be near 1 pixel using $0.01 \cdot \text{Sigmoid}(s) + 2.5 \cdot (1-\text{Sigmoid}(s))$ where $s$ is the output scale without activation, similar to \cite{xu2024grm}.

\subsection{Textured Mesh Extraction} 
From the generated multi-view outputs, we utilize RGB, depth, and opacity from Gaussian features for textured mesh extraction. First, we find the gradients of our generated depth map provides good approximation of surface normals. In addition, we mask out pixels with corresponding opacity values smaller than 0.1 and back-project the masked RGB/depth/normal pixels into 3D to create an oriented colored point cloud. We apply screened Poisson surface reconstruction \cite{kazhdan2013screened} to extract the mesh. Laplacian smoothing is applied to smooth the stair-case appearance. We generate texture coordinate using xatlas \cite{xatlas}, then assign color values on texture map by projecting color values from point cloud. Because our method produces pixel-perfect depth maps, our mesh extraction from point cloud (3D Gaussians) is accomplished without the need for any complex neural optimization processes, unlike methods such as LGM \cite{tang2024lgm}.

\subsection{Implementation Details}
We fine-tune the U-Net from Stable Diffusion Image Variation \cite{lambda_stable_diffusion, rombach2022high}. We initialize the depth branch of the U-Net with the weights of the first DownBlock and last UpBlock of the pre-trained U-Net and fine-tune all parameters of the U-Net. To enable high-resolution training with faster convergence, we first fine-tune U-Net on 256x256 resolution with a batch size of 512 for 30K iterations, and then fine-tune it on 512x512 resolution with a batch size of 96 for 100K iterations.

We fine-tune the VAE decoder together with our epipolar attention design from Stable Diffusion \cite{rombach2022high}. The fine-tuning is done on 512x512 resolution with a batch size of 8 for 90K iterations. We utilize a combination of regression loss and rendering loss for decoder training. After decoding multi-view outputs, we compute Mean Square Error (MSE) loss and LPIPS loss \cite{zhang2018unreasonable} for decoded RGB images and $\mathcal{L}_{L_1}$ loss and gradient matching loss $\mathcal{L}_{\text{gm}}$ \cite{ranftl2020towards} for decoded depth maps. We also back-project depths and Gaussian features to 3D and render the Gaussians via differentiable Gaussian splatting \cite{kerbl20233d}. We randomly render 10 novel views and compute MSE loss and LPIPS loss for rendered RGB images and MSE loss for rendered alpha images. The overall loss function is: 
\begin{equation}
\begin{aligned}
\mathcal{L} &= \mathcal{L}_{\text{Reg}} + \mathcal{L}_{\text{NVS}}, \\
\mathcal{L}_{\text{Reg}} &= \mathcal{L}_{\text{MSE}}^{\text{rgb}} + \lambda_{\text{LPIPS}} \mathcal{L}_{\text{LPIPS}}^{\text{rgb}} + \mathcal{L}_{L_1}^{\text{dep}} + \lambda_{\text{gm}} \mathcal{L}_{\text{gm}}^{\text{dep}}, \\
\mathcal{L}_{\text{NVS}} &= \mathcal{L}_{\text{MSE}}^{\text{rgb - nvs}} + \lambda_{\text{LPIPS}} \mathcal{L}_{\text{LPIPS}}^{\text{rgb - nvs}} + \mathcal{L}_{\text{MSE}}^{\text{alpha - nvs}},
\label{eq: loss2}
\end{aligned}
\end{equation}
where $\mathcal{L}^{\text{rgb}}$ and $\mathcal{L}^{\text{dep}}$ stand for losses over RGB and depth images across the six views, whereas $\mathcal{L}^{\text{rgb - nvs}}$, $\mathcal{L}^{\text{dep - nvs}}$ and $\mathcal{L}^{\text{alpha - nvs}}$ denote losses over synthesized novel views of RGB, depth and alpha images respectively.
We set $\lambda_{\text{LPIPS}}$ as 0.5 and $\lambda_{\text{gm}}$ as 2.


For both U-Net and decoder training, we use a learning rate of 1e-4. Our U-Net is conditioned by the CLIP embedding \cite{radford2021learning} of the input image via cross attention. The noisy RGB and depth latents are channel-concatenated with the input latent and then sent to U-Net for denoising \cite{liu2023zero}. We learn camera embeddings, transform them with an MLP, and add them to U-Net's timestep embeddings \cite{liu2023zero, long2024wonder3d, shi2023mvdream}. We utilize Xformers \cite{lefaudeux2022xformers} and FlashAttention \cite{dao2022flashattention, dao2023flashattention2} in U-Net and decoder training to enable fast and memory-efficient attention. For latent diffusion inference, we use a guidance scale of 3 and the number of diffusion steps is set to 50 using the DDIM \cite{song2020denoising} scheduler. Our image camera view follows Wonder3D's input view related system.

 \begin{figure*}[!htb]
\begin{center}
\centerline{\includegraphics[width=1.0\textwidth]{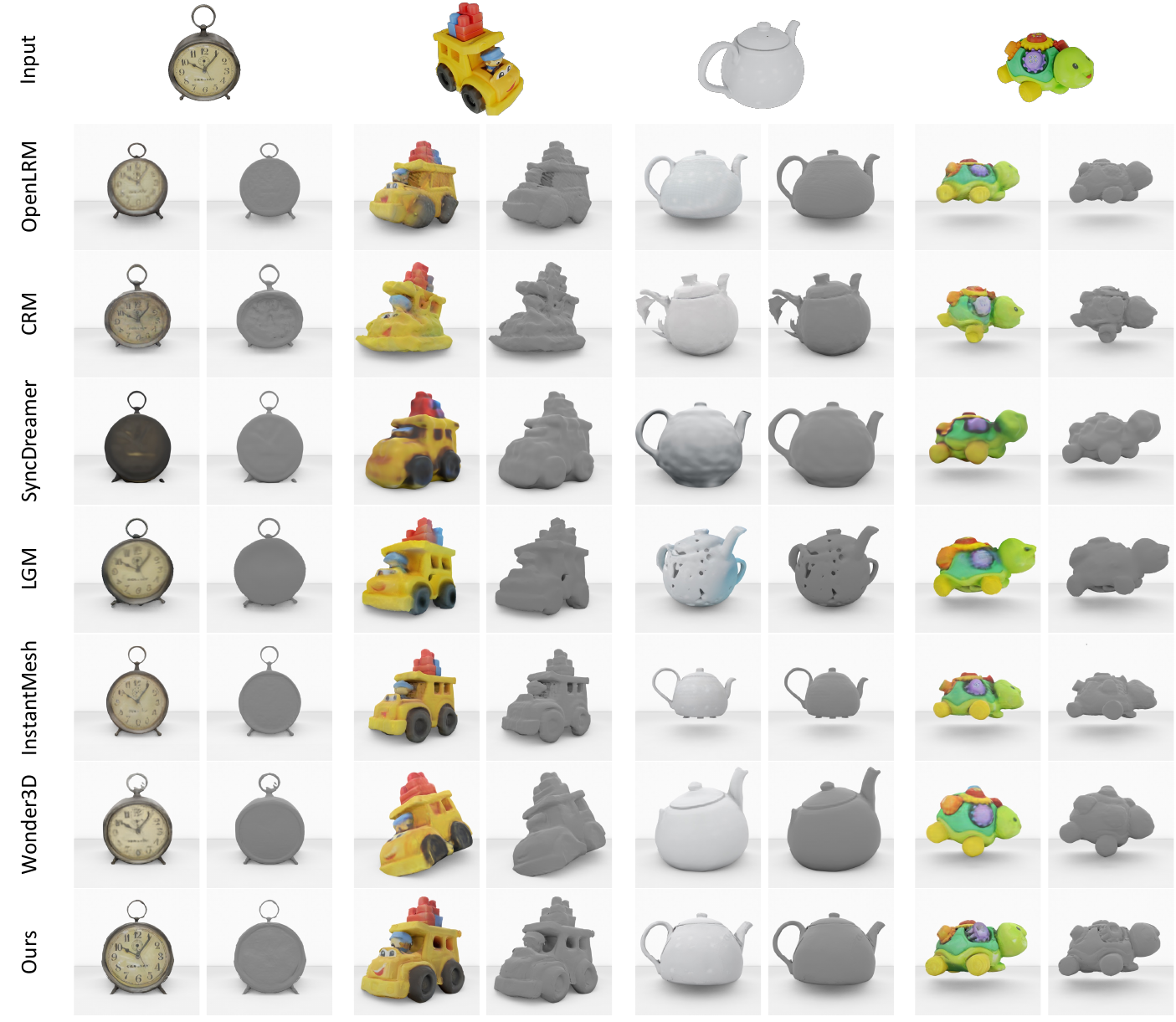}}
\end{center}
\vspace{-25pt}
\caption{\textbf{Qualitative results on GSO dataset.} We visualize the input single image and the resulting 3D mesh (with and without textures) for our method and baselines. Our approach achieves higher mesh quality in terms of both geometry and texture.}
\label{fig: compare}
\end{figure*}

\section{Experiments}
\label{sec:experiments}

\begin{table*}[!htb]
\centering
\begin{tabular}{lccccccc}
\toprule
Method   & Chamfer Dist.$\downarrow$ & Volume IoU$\uparrow$ & Depth Error$\downarrow$ & PSNR$\uparrow$ & SSIM$\uparrow$ & LPIPS$\downarrow$ & Time $\downarrow$  \\
\midrule
Realfusion \cite{melas2023realfusion}  & 0.1015 & 0.2882 & 0.394         & 12.44 & 0.764 & 0.373 & $\sim$1 hour  \\
Zero123 \cite{liu2023zero}     & 0.0627 & 0.4451 & 0.327         & 14.90 & 0.808 & 0.296 & $\sim$30 mins \\
Magic123 \cite{qian2023magic123}     & 0.0564 & 0.3988 & 0.282         & 10.50 & 0.770 & 0.386 & $\sim$1 hour  \\
\hline
Wonder3D  \cite{long2024wonder3d}    & 0.0236 & 0.6731 & 0.134         & 15.21 & 0.824 & 0.269 & 2-3 mins      \\
SyncDreamer \cite{liu2023syncdreamer} & 0.0234 & 0.6464 & 0.134         & 15.92 & 0.833 & 0.202 & 5-10 mins     \\
\hline
Point-E \cite{nichol2022point}    & 0.0520 & 0.2445 & 0.308         & 13.73 & 0.807 & 0.314 & 1-2 mins      \\
Shap-E   \cite{jun2023shap}     & 0.0438 & 0.3430 & 0.223         & 12.67 & 0.793 & 0.318 & 8-20 s        \\
LGM \cite{tang2024lgm}          & 0.0396 & 0.4538 & 0.210         & 14.09 & 0.833 & 0.328 & $\sim$1 min   \\
CRM \cite{wang2024crm}         & 0.0334 & 0.5594 & 0.173         & 14.02 & 0.835 & 0.309 & $\sim$30 s    \\
One-2-3-45 \cite{liu2024one}      & 0.0282 & 0.6131 & 0.143         & 16.51 & 0.838 & 0.217 & $\sim$45s     \\
InstantMesh \cite{xu2024instantmesh} & 0.0264 & 0.6584 & 0.143         & 16.51 & 0.842 & 0.205 & $\sim$30s     \\
OpenLRM \cite{he2023openlrm}     & 0.0186 & 0.7054 & 0.108         & 14.62 & 0.844 & 0.254 & $\sim$20s     \\
MVD-Fusion \cite{hu2024mvd} & 0.0362 &  ---  & --- &   ---  &  ---  &  ---  & $\sim$35s     \\
Ours & \textbf{0.0135} & \textbf{0.7339} & \textbf{0.073} & \textbf{17.85} & \textbf{0.851} & \textbf{0.159} & 15-25s \\
\bottomrule
\end{tabular}
\vspace{-5pt}
\caption{\textbf{Quantitative evaluation on the GSO dataset.} We report performance of baselines and our results on textured mesh generation. We classify the baselines into three categories: SDS optimization-based methods (first three rows), multi-view image (normal) generation-based optimization (lines 4-5), and direct feed-forward 3D generation methods (lines 6-14). We mark the best scoring methods with bold.}
\label{tab: gso_table}
\end{table*}



\noindent \textbf{Datasets.} For model training, we leverage the LVIS subset of the Objaverse dataset \cite{deitke2023objaverse}, similar to previous works \cite{liu2024one, long2024wonder3d}. This dataset contains around 46K objects in 1,156 categories. To generate ground-truth data, we normalize the objects to fit within a unit sphere. We use Blender \cite{blender} to render depth and RGB images from six orthogonal views: front, back, left, right, top and bottom. Additionally, random rotations are applied to objects to enrich the dataset. For evaluation, we follow prior research to use the Google Scanned Objects (GSO) dataset \cite{downs2022google}. We employ the same evaluation dataset as previous works \cite{liu2023syncdreamer, long2024wonder3d}, consisting of 30 common objects used in daily life. For each object, we render an image to serve as the input for our evaluation process. We also evaluate our method using additional object images collected by other methods \cite{long2024wonder3d, melas2023realfusion}. The image resolution is set to 512×512 for both training and evaluation.

\begin{figure*}[!htb]
\begin{center}
\centerline{\includegraphics[width=1\textwidth]{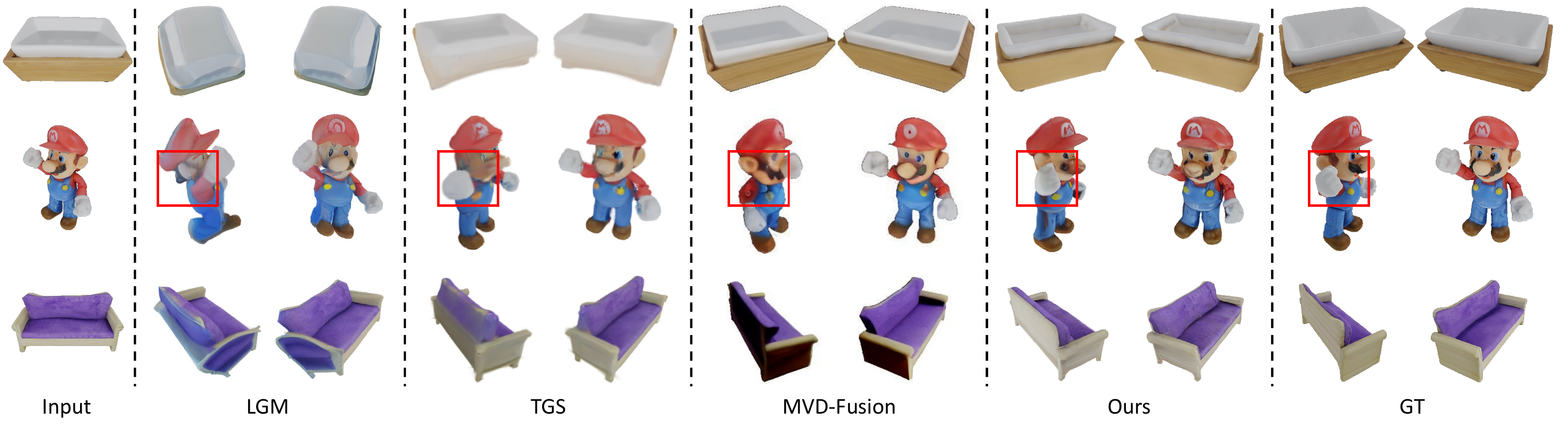}}
\end{center}
\vspace{-30pt}
\caption{\textbf{Qualitative comparisons for novel view synthesis}. Our generated 3D Gaussians deliver higher visual quality and capture more intricate details.}
\label{fig: compare_nvs}
\vspace{-5pt}
\end{figure*}

\noindent \textbf{Metrics.}
We follow the standard evaluation protocol in \cite{wang2024crm, liu2023syncdreamer, long2024wonder3d} for single-image 3D reconstruction. We report the Chamfer distance (CD) and the volume IoU between the reconstructed mesh and the ground-truth mesh. We also report depth map error (absolute distance) from mesh rendering as a measure of surface error. In experiment, we process each method's predictions using scale adaptive ICP \cite{sahilliouglu2021scale} to refine the alignment of the reconstructed mesh to ground truth. To measure texture quality, we render 512x512 images from the output mesh and the ground-truth mesh using 36 fixed camera views per object and report the average PSNR, SSIM \cite{wang2004image} and LPIPS \cite{zhang2018unreasonable}.

For novel view synthesis, we directly render 36 images via 3D Gaussian splatting from our output surface-aligned Gaussians and report PSNR, SSIM, and LPIPS metrics, following \cite{kerbl20233d}.


\noindent \textbf{Baselines.} We compare to state-of-the-art single image-to-3D methods including Zero123 \cite{liu2023zero}, RealFusion \cite{melas2023realfusion}, Magic123 \cite{qian2023magic123}, SyncDreamer \cite{liu2023syncdreamer}, Wonder3D \cite{long2024wonder3d}, Point-E \cite{nichol2022point}, Shap-E \cite{jun2023shap}, One-2-3-45 \cite{liu2024one}, CRM \cite{wang2024crm}, InstantMesh \cite{xu2024instantmesh}, OpenLRM \cite{he2023openlrm}, LGM \cite{tang2024lgm}, TGS \cite{zou2024triplane} and MVD-Fusion \cite{hu2024mvd}. We use their official implementations, except for LRM \cite{hong2023lrm} which only has open-sourced implementation OpenLRM \cite{he2023openlrm} available. For MVD-Fusion, since it produces coarse point clouds from low-resolution depth maps (32x32) and the textured mesh generation is not available, we only report its Chamfer distance for 3D reconstruction, following the original paper.

\subsection{Comparisons with SOTA Methods}

\begin{table}[!tb]
\vspace{-15pt}
\centering
\resizebox{0.85\columnwidth}{!}{
\begin{tabular}{lccc}
\toprule
Method   & PSNR$\uparrow$ & SSIM$\uparrow$ & LPIPS$\downarrow$ \\
\midrule
LGM \cite{tang2024lgm}       & 14.3465          & 0.8191          & 0.2991   \\
MVD-Fusion \cite{hu2024mvd}  & 16.5586          & 0.8314          & 0.2071     \\
TGS \cite{zou2024triplane}   & 17.5151         & 0.8612          & 0.2234   \\
Ours        & \textbf{18.1698} & \textbf{0.8621} & \textbf{0.1586} \\
\bottomrule
\end{tabular}}
\vspace{-7pt}
\caption{\textbf{Quantitative comparison for novel view synthesis on the GSO dataset}. We report performance of baselines and our results and mark the best scoring methods with bold.}
\label{tab: gso_table2}
\vspace{-4pt}
\end{table}

\noindent \textbf{3D Reconstruction.}
Our approach demonstrates superior performance compared to state-of-the-art methods in single-image 3D reconstruction, both qualitatively as shown in \cref{fig: compare} and quantitatively as shown in \cref{tab: gso_table}. Compared to other baselines, our method consistently generates meshes with better texture fidelity and geometric accuracy. Notably, our approach performs exceptionally well in capturing intricate details because we generate high-resolution depth maps with multi-view consistency at the pixel level. Furthermore, our method also achieves competitive generation speed compared to other feed-forward methods, second only to Shap-E but showing much better geometric quality.

\noindent \textbf{Novel View Synthesis.} Our method also outperforms other baselines in novel view synthesis as shown in~\cref{tab: gso_table2} and~\cref{fig: compare_nvs}. Compared to LGM \cite{tang2024lgm} and TGS \cite{zou2024triplane}, which also produce 3D Gaussians, our renderings exhibit superior visual quality. We observe that MVD-Fusion \cite{hu2024mvd}, despite utilizing depth-aware 3D attention, still results in multi-view inconsistencies. In contrast, our method generates dense, surface-aligned Gaussians, ensuring multi-view consistency and yielding detailed and accurate renderings.

\subsection{Ablation Study}

\begin{table}[!tb]
\centering
\resizebox{0.95\columnwidth}{!}{
\begin{tabular}{lcccc}
\toprule
Method    & CD$\downarrow$  & IoU$\uparrow$   & PSNR$\uparrow$ & LPIPS$\downarrow$ \\
\midrule
Full     & \textbf{0.0135} & \textbf{0.7339} & \textbf{17.85} & \textbf{0.159} \\
w/o Epipolar Attn. & 0.0323 &	0.4518 & 15.12 &	0.262  \\
w/o Depth Latent   & 0.0249 &	0.6010 & 15.40 &	0.238  \\
w/o NVS loss & 0.0136 & 0.7286 &	17.77 &	0.166  \\
\bottomrule
\end{tabular}}
\vspace{-7pt}
\caption{\textbf{Quantitative ablations on our design choices.} We mark the best scoring methods with bold.}
\label{tab: ablation}
\vspace{-5pt}
\end{table}

\begin{table}[!htb]
\centering
\resizebox{\columnwidth}{!}{
\begin{tabular}{lccccc}
\toprule
Method & CD$\downarrow$  & IoU$\uparrow$ & PSNR$\uparrow$  & GPU$\downarrow$ & Time$\downarrow$   \\
\midrule	
Ours &  \textbf{0.0152}  & \textbf{0.7303} & 17.19 & 9 GB & 20 s  \\
w/o Branch & 0.0166  & 0.6716  & \textbf{17.20} & 11 GB & 24 s   \\
\bottomrule
\end{tabular}}
\vspace{-8pt}
\caption{\textbf{Ablations on our branched U-Net design}. We replace our branched U-Net design with domain-switch \cite{long2024wonder3d} for cross-domain latent denoising. Note that the study was performed at 256x256 resolution due to compute resource constraints.}
\label{tab: ablation_unet}
\vspace{-2pt}
\end{table}

\begin{table}[!htb]
\centering
\resizebox{\columnwidth}{!}{
\begin{tabular}{lccccc}
\toprule
Method & CD$\downarrow$  & IoU$\uparrow$   & PSNR$\uparrow$ & LPIPS$\downarrow$ & Time$\downarrow$ \\
\midrule	
Wonder3D & 0.0236 & 0.6731 & 15.21 & 0.269 & 3min \\
Ours$_{Normal}$ & 0.0194 &	0.6930	& 15.43 &0.261 & 3min \\
Ours  &  \textbf{0.0152}  & \textbf{0.7303} & \textbf{17.19} & \textbf{0.171} & 20s  \\
\bottomrule
\end{tabular}}
\vspace{-10pt}
\caption{\textbf{Depth maps vs. normal maps as the representation.} Note that the study was performed at 256x256 resolution due to compute resource constraints.}
\label{tab: abl_normal}
\vspace{-5pt}
\end{table}

\begin{figure}[!tb]
\begin{center}
\vspace{-10pt}
\centerline{\includegraphics[width=0.5\textwidth]{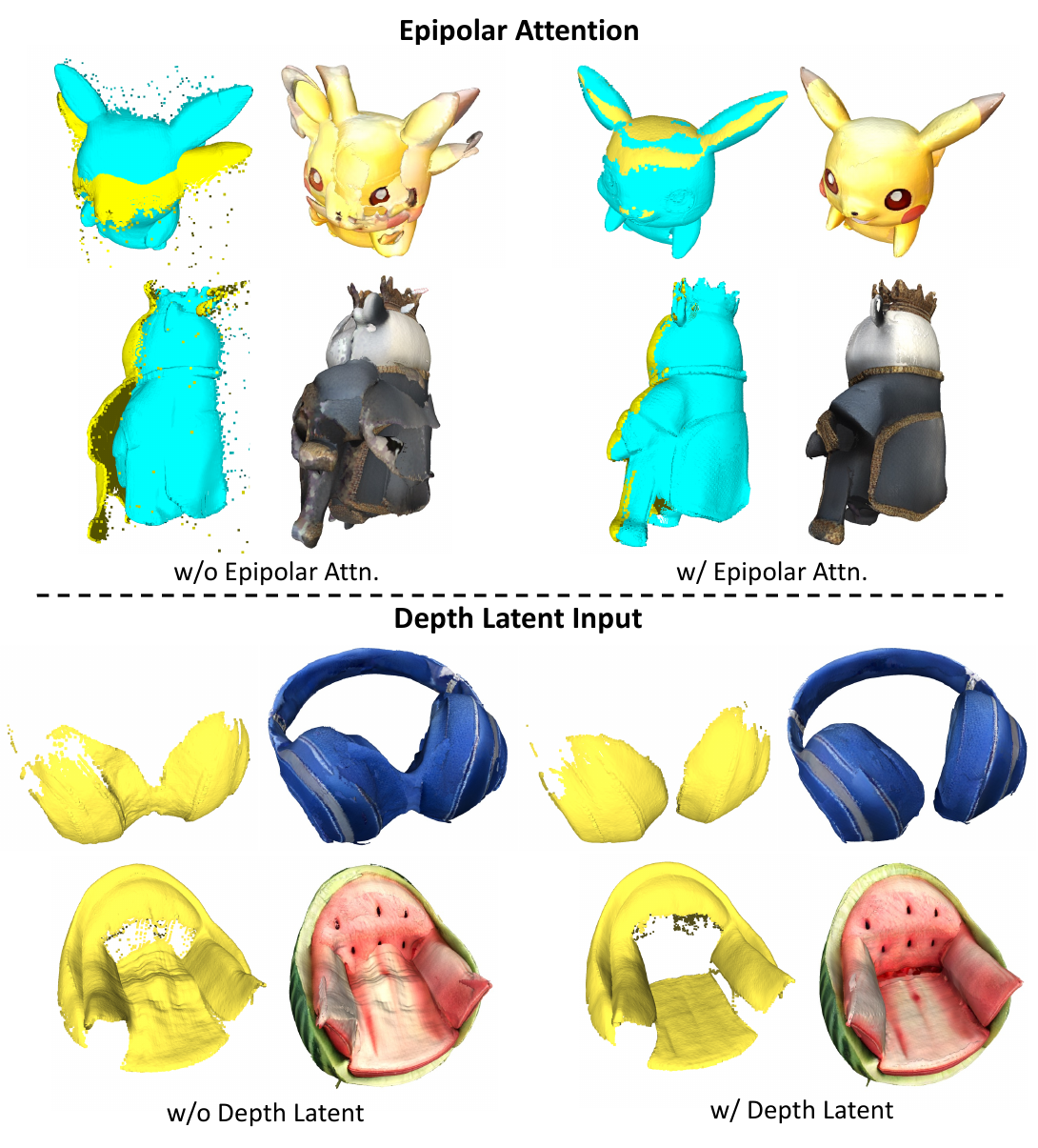}}
\end{center}
\vspace{-30pt}
\caption{\textbf{Qualitative ablations on the GSO dataset.} We show back-projected depth points for 1 (yellow) or 2 views (yellow, cyan) and the final mesh.}
\label{fig: abl}
\end{figure}

We conduct ablation studies on the GSO dataset, as shown in \cref{tab: ablation} and \cref{fig: abl}. First, we show that removing epipolar attention in the depth decoder leads to a major drop of all quantitative metrics, due to the severe inconsistent multi-view depth maps and distorted mesh extraction, highlighting the indispensable role of epipolar attention. Second, without learning to generate the depth latent and provide it as additional input for latent decoding, \ie only generating RGB latent and feed it to the decoder, we observe that the predicted depths are of lower quality and are inaccurate. In addition, to validate our branched U-Net design that performs simultaneous multi-domain latent denoising, we experiment replacing it with domain-switch technique from Wonder3D \cite{long2024wonder3d} which performs successive multi-domain denoising, as shown in \cref{tab: ablation_unet}. Our branched U-Net achieves 20\% faster in inference and uses 18\% less GPU memory, with no degradation in quality. 

\textbf{Is depth representation better than normal maps within the same framework?}
Our approach outperforms both in terms of generation quality and efficiency compared to predicting normal maps using our framework or Wonder3D (\cref{tab: abl_normal}), which also predicts multi-view RGBD for 3D reconstruction. Normals do not represent depth discontinuities and require optimization to reconstruct the geometry, which often results in incorrect or blurry geometry.
 \section{Conclusion}
We present an approach to directly predict explicit surface geometry and texture for single-image 3D reconstruction. 
Experiments show that our method significantly improves the speed and quality of 3D reconstructions compared to other benchmarks. 

\noindent\textbf{Acknowledgment.} This research was supported by funding from Amazon, the NASA Biodiversity Program (Award 80NSSC21K1027), and the National Science Foundation (NSF) Grant IIS-2212046.  
\clearpage
{
    \small
    \bibliographystyle{ieeenat_fullname}
    \bibliography{main}
}
\appendix 
\clearpage
\setcounter{page}{1}
\maketitlesupplementary

In \cref{sec: supp_results}, we provide an example of our generated multi-view outputs, additional comparisons on single-image 3D reconstruction, an analysis on the number of views, and a comparison with monocular depth estimators. In \cref{sec: supp_arch}, we provide more details on our model architecture. In \cref{sec: supp_settings}, we describe our experimental settings in more detail. In \cref{sec: text_mesh}, we show how to extend our approach to obtain rigged and posed meshes. In \cref{sec: limit}, we discuss the limitations of our method. We also include a supplementary video that compares our method's results against baseline methods and shows additional results of our approach.

\section{Additional Results}
\label{sec: supp_results}

\begin{figure}[!h]
\begin{center}
\centerline{\includegraphics[width=0.5\textwidth]{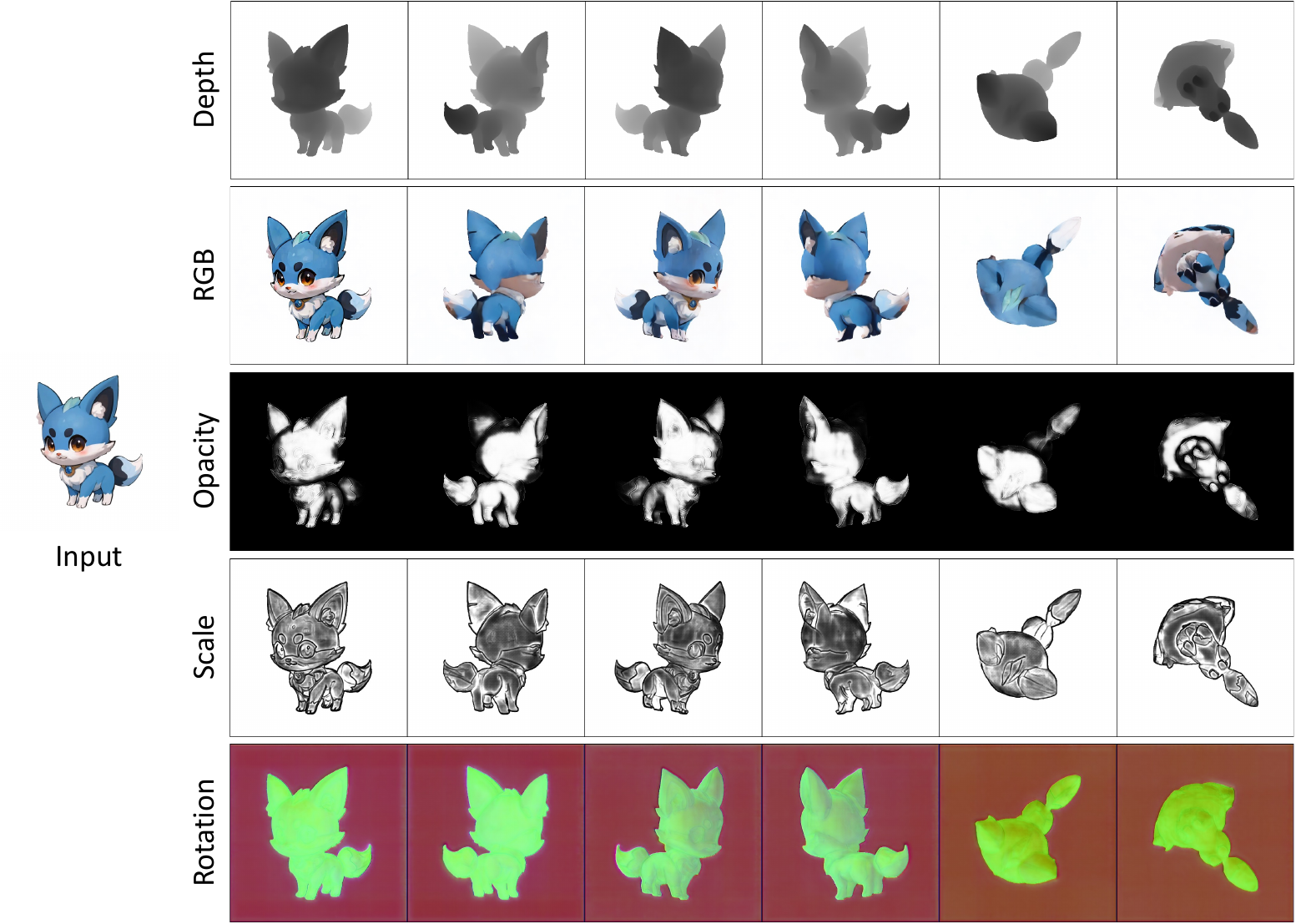}}
\end{center}
\vspace{-25pt}
\caption{An example of our generated multi-view depth, RGB, and Gaussian feature images. For rotation of quaternion $\mathbf{q} \in \mathbb{R}^4$, we visualize its last three channels.}
\label{fig: vis_output}
\vspace{-5pt}
\end{figure}

\noindent \textbf{Our Multi-view Outputs.}
Given an input image, our method generates depth map along with RGB and Gaussian feature maps in six orthographic views (relative camera poses from the front, back, left, right, top, and bottom). In \cref{fig: vis_output}, we present an example of our multi-view predictions.

\begin{figure*}[!h]
\begin{center}
\centerline{\includegraphics[width=1.0\textwidth]{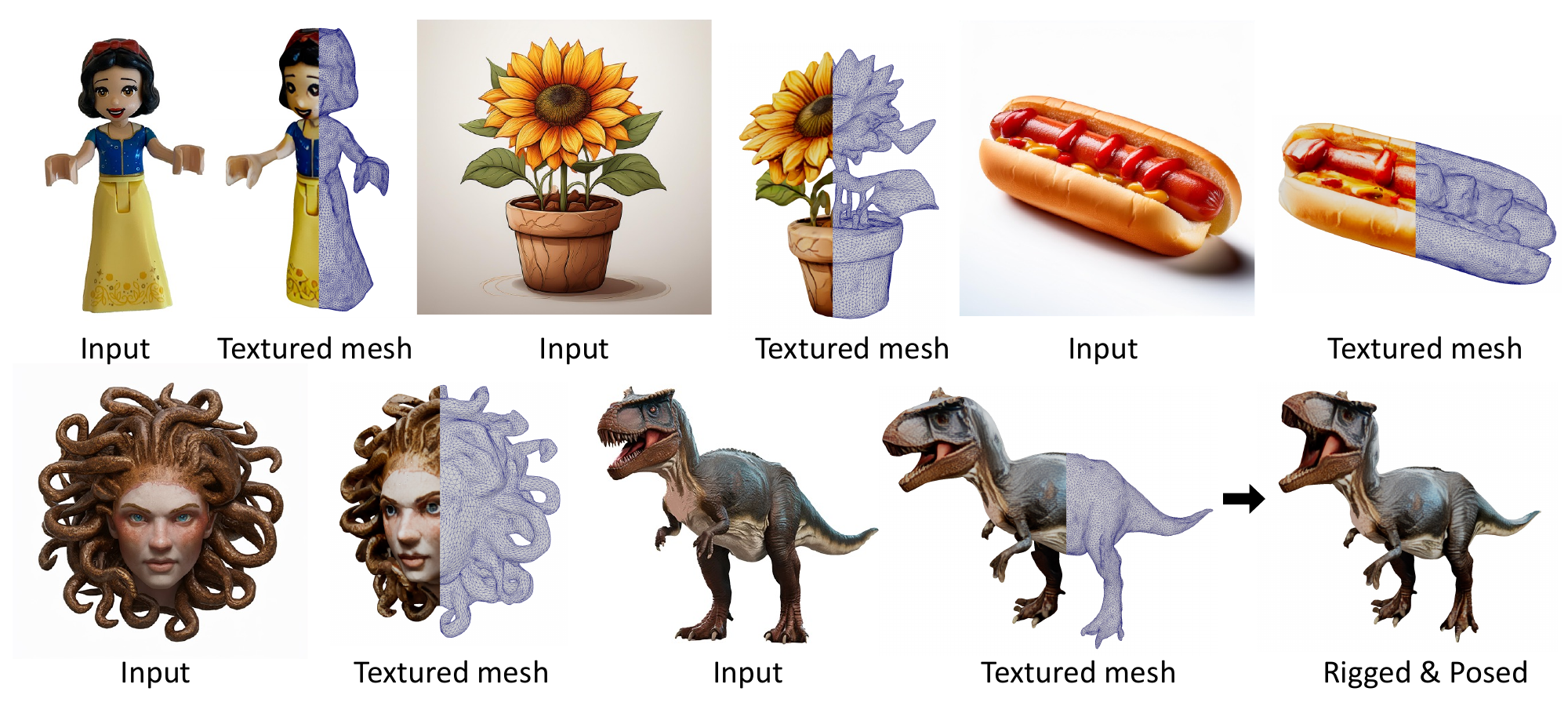}}
\end{center}
\vspace{-30pt}
\caption{Refined coarse quality triangulated meshes and a rigged and re-posed example (bottom right).}
\label{fig: tex_mesh}
\end{figure*}

\begin{figure*}[!h]
\begin{center}
\centerline{\includegraphics[width=0.97\textwidth]{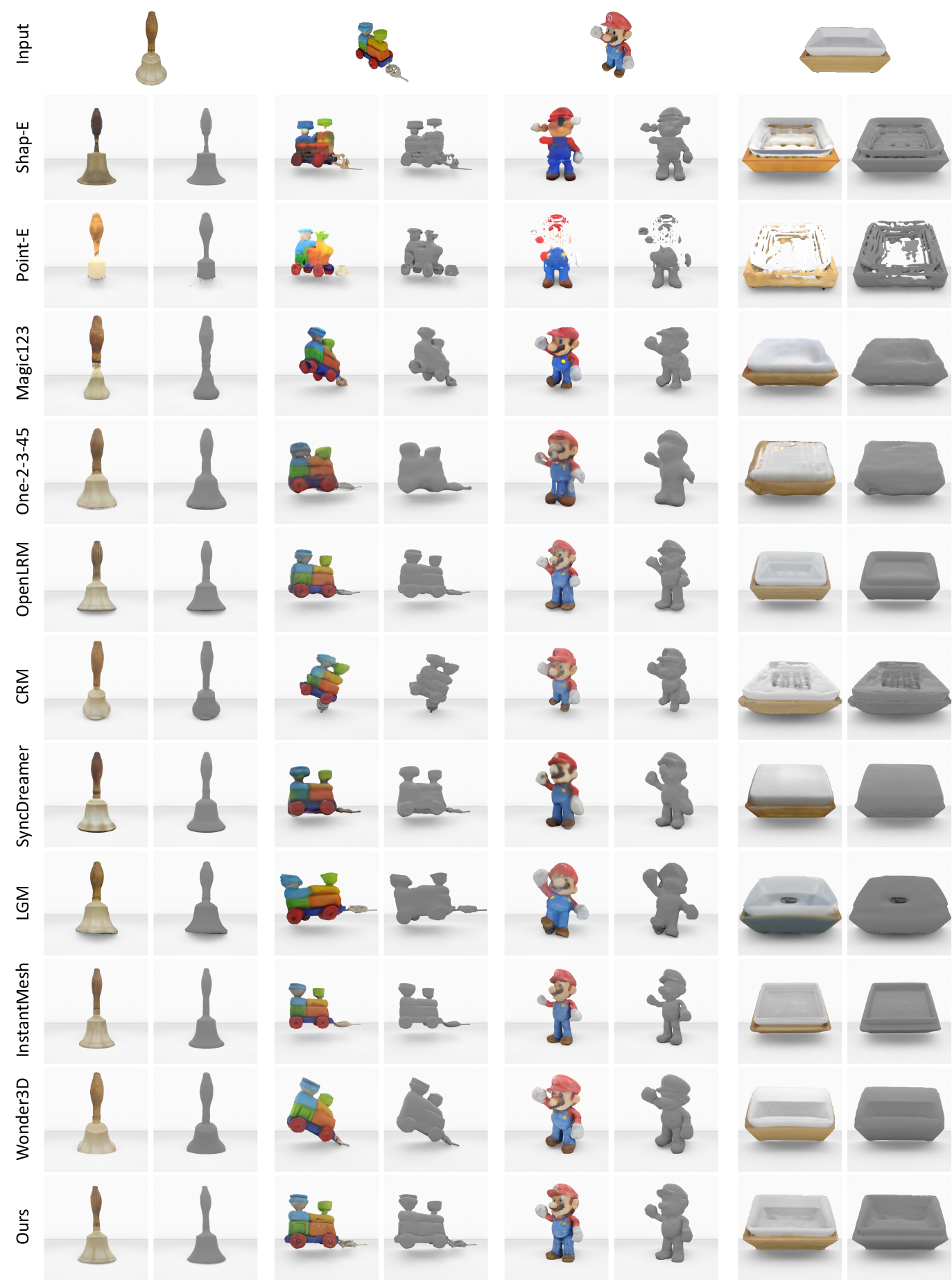}}
\end{center}
\vspace{-30pt}
\caption{Additional comparisons on generated textured meshes.}
\label{fig: supp_compare}
\end{figure*}

\noindent \textbf{Additional Comparison on Single-image 3D Reconstruction.} 
In \cref{fig: supp_compare}, we provide additional qualitative comparisons with other methods on generated textured meshes on the GSO dataset \cite{downs2022google}. Our results appear to have higher quality and better details in both geometry and texture.

\noindent \textbf{Number of Views.} 
We conduct an empirical study on the relationship between the number of views for RGB and depth images used to reconstruct a 3D object and the quality of its reconstruction.
We use Objaverse dataset \cite{deitke2023objaverse} for this study, which contains a wide range of 800K objects.
We randomly sample 1,000 objects from 18 high-level categories on Objaverse dataset. We make sure that the number of objects we sample from each category matches the original percentage of that category. For the sampled objects, we attempt to reconstruct textured mesh using 4, 6, 8, or 14 views of RGB and depth images, and report the quality of the reconstruction. The view names in orders are front, back, left, and right, top, bottom, right-top-front, right-top-back, right-bottom-front, right-bottom-back, left-top-front, left-top-back, left-bottom-front, and left-bottom-back. We use screened Poisson surface reconstruction \cite{kazhdan2013screened} to obtain the mesh. 
As shown in \cref{tab: nviews}, increasing the number of views consistently improves reconstruction quality. The effect diminishes after 6 views, where the improvement from 4 to 6 views is significant, but further gains from 6 to 8 or 14 views are relatively smaller.

\begin{table}[!htb]
\centering
\resizebox{1\columnwidth}{!}{
\begin{tabular}{lccccc}
\toprule
Views & CD$\downarrow$  & IoU$\uparrow$   & PSNR$\uparrow$ & SSIM$\uparrow$ & LPIPS$\downarrow$ \\
\midrule
4  & 0.0078 & 0.7468  &  23.75 & 0.926 & 0.060 \\
6  & 0.0070 & 0.7661  &  25.44 & 0.938 & 0.049 \\
8  & 0.0068 & 0.7687 &   25.67 & 0.939 & 0.046 \\
14 & 0.0062 & 0.7780  &  26.37 & 0.946 & 0.041 \\
\bottomrule
\end{tabular}}
\caption{Comparison of reconstruction quality for different numbers of views on Objaverse dataset.}
\label{tab: nviews}
\end{table}

\noindent \textbf{Comparison with monocular depth estimators.}
We compare our method with other single-image depth estimation methods in \cref{tab: mono_depth}. This study is conducted on 3D objects using the same GSO \cite{downs2022google} evaluation dataset as in the main text. For a fair comparison, we use our predicted depth map for the front (input) view as our single-view depth estimation result. Following prior works \cite{ranftl2020towards, ranftl2021vision}, we evaluate and report the mean absolute value of the relative error in depth space (AbsRel).

\begin{table}[!htb]
\centering
\resizebox{1\columnwidth}{!}{
\begin{tabular}{lcccc}
\toprule
 & MiDaS \cite{ranftl2020towards} & DPT \cite{ranftl2021vision} & Omnidata \cite{eftekhar2021omnidata} & Ours \\
\midrule
AbsRel (\%)$\downarrow$  & 17.3 & 13.5 & 12.6 & \textbf{6.37} \\
\bottomrule
\end{tabular}}
\caption{Comparisons on single-image depth estimation.}
\label{tab: mono_depth}
\end{table}

\begin{table*}[!htb]
\centering
\begin{tabular}{l|c|c}
\toprule
U-Net                      & Stable Diffusion     & Ours                   \\
\midrule
Input/Output               & B, 4, H/8, W/8       & B*6, 8, H/8, W/8     \\
\hline
\multirow{4}{*}{Down Blocks} & CrossAttnDownBlock2D & CrossAttnDownBlockMV2D x 2 (RGB, Depth) \\
                           & CrossAttnDownBlock2D & CrossAttnDownBlockMV2D \\
                           & CrossAttnDownBlock2D & CrossAttnDownBlockMV2D \\
                           & DownBlock2D          & DownBlock2D            \\
\hline
Middle Block & UNetMidBlockMV2DCrossAttn & UNetMidBlockMV2DCrossAttn \\
\hline
\multirow{4}{*}{Up Blocks} & UpBlock2D            & UpBlock2D              \\
                           & CrossAttnUpBlock2D   & CrossAttnUpBlockMV2D   \\
                           & CrossAttnUpBlock2D   & CrossAttnUpBlockMV2D   \\
                             & CrossAttnUpBlock2D   & CrossAttnUpBlockMV2D x 2 (RGB, Depth)   \\
\bottomrule
\end{tabular}
\caption{Comparison between our U-Net and Stable Diffusion U-Net \cite{rombach2022high}.}
\label{tab: supp_unet}
\end{table*}

\begin{table*}[!htb]
\centering
\begin{tabular}{l|c|c}
\toprule
Decoder        & Stable Diffusion & Ours                            \\
\hline
Input                   & B, 4, H/8, W/8   & B*6, 8, H/8, W/8                \\
\hline
Output                  & B, 3, H, W       & B*6, 12, H, W                   \\
\hline
\multirow{4}{*}{Blocks} & UpDecoderBlock2D & (Epipolar) AttnUpDecoderBlock2D \\
                        & UpDecoderBlock2D & (Epipolar) AttnUpDecoderBlock2D \\
                        & UpDecoderBlock2D & (Epipolar) AttnUpDecoderBlock2D \\
                        & UpDecoderBlock2D & (Epipolar) AttnUpDecoderBlock2D \\
\bottomrule
\end{tabular}
\caption{Comparison between our decoder and the VAE decoder in Stable Diffusion \cite{rombach2022high}.}
\label{tab: supp_decoder}
\end{table*}

\section{Additional Model Details}
\label{sec: supp_arch}

\noindent \textbf{Network Architectures.} 
As mentioned in the main text, we add a depth branch to the Stable Diffusion U-Net and incorporate epipolar attention into the Stable Diffusion VAE decoder. We compare our U-Net and decoder architectures with those in Stable Diffusion in \cref{tab: supp_unet} and \cref{tab: supp_decoder}, respectively.

\noindent \textbf{Training Configuration.} The following training configurations are applied to the fine-tuning of both the U-Net and the latent decoder. 
\begin{verbatim}
training config:
    optimizer="adam",
    adam_beta1=0.9,
    adam_beta2=0.999,
    adam_eps=1e-8,
    learning_rate=1e-4,
    weight_decay=0.01,
    gradient_clip_norm=1.0,
    ema_decay=0.9999,
    mixed_precision_training=bf16
\end{verbatim}

\section{Additional Experimental Settings}
\label{sec: supp_settings}

\noindent \textbf{Compensating Global Similarity Using Iterative Closest Point.}
As we perform 3D reconstruction from single view images, global scale and rigid pose of the underlying objects cannot be resolved uniquely, introducing a global similarity ambiguity. Therefore, before applying geometric metrics such as Chamfer Distance and Volume IoU, we perform similarity alignment of our estimated shape with the ground-truth shape following standard practice of prior works (as listed in Table 1 from the main document). Specifically, We extended scale adaptive ICP \cite{sahilliouglu2021scale} to identify optimal scale factors along each coordinate axes, in addition to its original uniform scale and translation.

\section{Application: Refining Extracted Textured Mesh for Deformations}
\label{sec: text_mesh}
Here we show how our approach can be extended to obtain rigged and posed meshes. Our initial mesh is reconstructed by screened Poisson surface reconstruction \cite{kazhdan2013screened}, which typically consists of millions of uneven triangles with possible unnecessary outlier pieces. To improve the quality of the triangles and reduce their number for better rigging and posing, we perform additional refinement steps.
First, we remove any small pieces that are disconnected from the main component. Next, we generate a cage mesh that encapsulates the original mesh, following the method described in \cite{Sellan:Breaking:2022}. We then perform Non-rigid ICP \cite{amberg2007optimal} to register the cage mesh to the original mesh. The registered cage mesh, now aligned with the original mesh, allows us to control the number and quality of the triangles, resulting in the final refined output mesh. In \cref{fig: tex_mesh}, we provide examples of refined meshes, including one that has been rigged and re-posed.

\begin{table}[!h]
\centering
\resizebox{1\columnwidth}{!}{
\begin{tabular}{lccccc}
\toprule
Method & CD$\downarrow$  & IoU$\uparrow$   & PSNR$\uparrow$ & SSIM$\uparrow$ & LPIPS$\downarrow$ \\
\midrule
Ours  & 0.0135 & 0.7339 & 17.85 & 0.851 & 0.159 \\
Ours-Persp.  & 0.0138	& 0.7272 & 17.70 & 0.848 & 0.159 \\
\bottomrule
\end{tabular}}
\caption{Comparison between our model and a variant that is trained with perspective images as the input.}
\label{tab: persp_model}
\end{table}

\section{Limitations}
\label{sec: limit}

One limitation of our approach is the assumption that the input images are orthogonal, which may lead to distortion in the generated results, even though we do not see many visual artifacts when using perspective images as input in inference.

We tried training the model using perspective images with fixed focal length, and obtained results similar to but slightly worse than our main model trained based on orthogonal views (\cref{tab: persp_model}). Also note that the model trained using perspective images is still specific to the camera type. Therefore, developing a model that can handle images from various camera types remains an open and interesting research direction \cite{li2024era3d}.

\end{document}